\documentclass{article}
\usepackage{spconf,amsmath,graphicx}
\usepackage{epstopdf}
\usepackage{epsfig}
\usepackage{amsfonts}
\usepackage{amssymb}
\newcommand{\argmin}{\operatornamewithlimits{argmin}}
\usepackage{amsthm}
\usepackage{url}
\newtheorem{thm}{Theorem}

\usepackage{amsfonts}
\usepackage{amssymb}
\usepackage{latexsym}
\usepackage{caption}
\usepackage{subcaption}

\title{Orthogonal Matrix Retrieval in Cryo-electron microscopy}
\name{Tejal Bhamre, Teng Zhang and Amit Singer 
\thanks{This research was partially supported by Award Number R01GM090200 from the NIGMS and Award Number N66001-13-1-4052 from DARPA.}}
\address{Princeton University, Program in Applied and Computational Mathematics, Princeton NJ, USA}
\begin{document}
%
\maketitle
\begin{abstract}
In single particle reconstruction (SPR) from cryo-electron microscopy (EM), the 3D structure of a molecule needs to be determined from its 2D projection images taken at unknown viewing directions. Zvi Kam showed already in 1980 that the autocorrelation function of the 3D molecule over the rotation group SO(3) can be estimated from 2D projection images whose viewing directions are uniformly distributed over the sphere. The autocorrelation function determines the expansion coefficients of the 3D molecule in spherical harmonics up to an orthogonal matrix of size $(2l+1)\times (2l+1)$ for each $l=0,1,2,\cdots$. In this paper we show how techniques for solving the phase retrieval problem in X-ray crystallography can be modified for the cryo-EM setup for retrieving the missing orthogonal matrices. Specifically, we present two new approaches that we term {\em Orthogonal Extension} and {\em Orthogonal Replacement}, in which the main algorithmic components are the singular value decomposition and semidefinite programming. We demonstrate the utility of these approaches through numerical experiments on simulated data.
\end{abstract}
\section{Introduction}
\label{sec:intro}

SPR from cryo-EM is an increasingly popular technique in structural biology for
determining 3D structures of macromolecular complexes that resist crystallization \cite{Frank1,resolution_revolution,Revol}.
In the basic setup of SPR, the data collected are 2D projection images of ideally assumed identical, but randomly oriented, copies of a macromolecule. In cryo-EM, the sample of molecules is rapidly frozen in a thin layer of vitreous ice, and maintained at liquid nitrogen temperature throughout the imaging process \cite{wang06}. The electron microscope provides a top view of the molecules in the form of a large image called a micrograph. The projections of the individual particles can be picked out from the micrograph, resulting in a set of projection images. Datasets typically range from $10^4$ to $10^5$ projection images whose size is roughly $100 \times 100$ pixels.

Mathematically, ignoring the effects of the microscope's contrast transfer function and noise, a 2D projection image $I:\mathbb{R}^2\to \mathbb{R}$ corresponding to rotation $R$ is given by the integral of the Coulomb potential $\phi : \mathbb{R}^3 \to \mathbb{R}$ that the molecule induces
\begin{equation}
\label{forward-model}
I(x,y) = \int_{-\infty}^\infty \phi(R^T r)\,dz,
\end{equation}
where $r = (x, y, z)^T$. The 3D reconstruction problem in cryo-EM is a non-linear inverse problem in which $\phi$ needs to be estimated from multiple noisy discretized projection images of the form (\ref{forward-model}) for which the rotations are unknown.

Radiation damage limits the maximum allowed electron dose. As a result, the acquired 2D projection images are extremely noisy with poor signal-to-noise ratio (SNR). Estimating $\phi$ and the unknown rotations at very low SNR is a major challenge.

The 3D reconstruction problem is typically solved by guessing an initial structure and then performing an iterative refinement procedure, where iterations alternate between estimating the rotations given a structure and estimating the structure given rotations \cite{Frank1, vanheel00, gridding}. When the particles
are too small and images too noisy, the final result of the refinement process
depends heavily on the choice of the initial model, which makes
it crucial to have a good initial model. If the molecule is known to have a
preferred orientation, then it is possible to
find an \textit{ab-initio} 3D structure using the random conical tilt
method ~\cite{Radermacher, Radermacher2}. There are two known approaches to ab initio estimation that do not involve tilting: the
method of moments~\cite{Goncharov,Salzman}, and common-lines based methods \cite{Goncharov1986,vanHeel1987,SS_11}. 

Using common-lines based approaches, ~\cite{Zhao} was able to obtain three-dimensional
ab-initio reconstructions from real microscope
images of large complexes that had undergone only rudimentary averaging.
However, 
researchers have so far been
unsuccessful in obtaining meaningful 3D ab-initio models directly from raw
images that have not been averaged, especially for small complexes.

We present here two new approaches for ab-initio modelling that are based on Kam's theory \cite{kam1980} and that can be regarded
as a generalization of the molecular
replacement method from X-ray crystallography to cryo-EM. The only requirement
for our methods to
succeed is that the number of collected images is large enough for accurate estimation of the covariance matrix of the 2D projection images.

\section{Kam's theory and the Orthogonal matrix retrieval problem}
Kam showed \cite{kam1980} using the Fourier projection slice theorem (see, e.g., \cite[p. 11]{Natr2001a}) that if the viewing
directions of the projection images are uniformly distributed over the sphere, then the
autocorrelation function of the 3D volume with itself over the rotation group
SO(3) can be directly computed from the
covariance matrix of the 2D images. Let $\hat \phi : \mathbb{R}^3 \to \mathbb{C}$ be the 3D Fourier
transform of $\phi$ and consider its expansion in spherical coordinates 
\begin{equation}
\hat{\phi}(k,\theta,\varphi) = \sum_{l=0}^{\infty} \sum_{m=-l}^{l} A_{lm}(k) Y_l^m
(\theta, \varphi)
\end{equation}
where $k$ is the radial frequency and $Y_l^m$ are the real spherical
harmonics. Kam showed that   
\begin{equation}
\label{Cl}
C_l(k_1,k_2) = \sum_{m=-l}^l A_{lm}(k_1)\overline{A_{lm}(k_2)}  
\end{equation}
can be estimated from the covariance matrix of the 2D projection images. For images sampled on a Cartesian grid, each matrix $C_l$ is of size $K\times K$, where $K$ is the maximum frequency (dictated by the experimental setting). In matrix notation, eq.(\ref{Cl}) can be rewritten as
\begin{equation}
C_l=A_l A_l^*, \label{eq:Cl}
\end{equation}
where $A_l$ is a matrix size
$K \times (2l+1)$ whose $m$'th column is $A_{lm}$. 
The factorization (\ref{eq:Cl}) of $C_l$, also known as the Cholesky decomposition, is not unique: If
$A_l$ satisfies \eqref{eq:Cl}, then $A_lU$ also satisfies \eqref{eq:Cl} for any $(2l+1) \times (2l+1)$ unitary matrix
$U$ (i.e., $UU^* = U^*U = I$). 

Since $\phi$, the electric potential induced by the molecule, is
real-valued, its Fourier transform $\hat{\phi}$ satisfies
$\hat{\phi}(r)=\overline{\hat{\phi}(-r)}$, or equivalently,
$\hat{\phi}(k,\theta,\varphi)=\overline{\hat{\phi}(k,\pi-\theta,\varphi+\pi)}$.
Together with properties of the real spherical harmonics, it follows that
$A_{lm}(k)$ (and therefore $A_l$) is real for even
$l$ and purely imaginary for odd $l$. Then $A_l$ is unique up to a
$(2l+1)\times (2l+1)$ orthogonal matrix $O_l\in \text{O}(2l+1)$, where 
\begin{equation}
\text{O}(d)=\{O\in \mathbb{R}^{d\times d} \,:\, OO^T=O^TO=I \}.
\end{equation}

Originally, $2l+1$ functions of
the radial frequency are required for each $l$ in order
to completely characterize $\phi$. With the additional knowledge of
$C_l$ the parameter space is reduced
to $\text{O}(2l+1)$. We refer to the problem of recovering the missing orthogonal matrices $O_0,O_1,O_2,\ldots$ as the {\em orthogonal matrix retrieval problem in cryo-EM}.  

\subsection{Analogy with X-ray crystallography}
The orthogonal matrix retrieval problem is akin to the phase retrieval problem in X-ray crystallography. In crystallography,  
the measured diffraction patterns contain information about the modulus of the 3D
Fourier transform of the structure but the phase information is missing and needs to be obtained by other means. Notice that in crystallography,
the particle's orientations are known but
the phases of the Fourier coefficient are missing, while in electron
microscopy, the projection images contain phase information but the orientations of the
particles are missing. Kam's theory converts the cryo-EM problem to one akin to the phase retrieval problem in crystallography. From a mathematical standpoint, the phase retrieval problem in crystallography is perhaps more challenging than the orthogonal matrix retrieval problem in cryo-EM, because in crystallography each Fourier coefficient is missing its phase, while in cryo-EM only a single orthogonal matrix is missing per several radial components.

\section{Orthogonal Extension (OE)}
A classical solution to the missing phase problem in crystallography is
molecular replacement, which relies upon the existence of a previously solved structure which is similar to the unknown structure from which the
diffraction data is obtained. The structure is then estimated using the Fourier magnitudes from the diffraction data with the phases from the homologous structure. We mimic this approach in cryo-EM, by grafting the
orthogonal matrices of the already resolved
similar structure onto the unknown structure. 

Let $\phi$ be the unknown structure, and suppose $\psi$ is a known homologous structure, whose 3D Fourier transform $\hat{\psi}$ has the following expansion in spherical harmonics 
\begin{equation}
\hat{\psi}(k,\theta,\varphi) = \sum_{l=0}^{\infty} \sum_{m=-l}^{l}
B_{lm}(k) Y_l^m (\theta, \varphi)
\end{equation} 

We can obtain the auto-correlation matrices $C_l$ from the cryo-EM images of the
unknown structure $\phi$ using Kam's method.
Let $F_l$ be any matrix satisfying $C_l= F_l F_l^*$, determined from the
Cholesky decomposition of $C_l$. Then 
\begin{equation}
A_l = F_l O_l
\end{equation}
where $O_l \in \text{O}(2l+1)$. Requiring $A_l \approx B_l$, in {\em orthogonal extension} we determine $O_l$ as the solution to the
least squares problem
\begin{equation}\label{ls}
O_l=\argmin_{O \in \text{O}(2l+1)}  \|F_l O - B_l\|_F^2,
\end{equation}
where $\|\cdot\|_F$ denotes the Frobenius norm. 

Although the orthogonal group is non-convex, there is a closed form solution to (\ref{ls}) (see, e.g., \cite{keller}) given by
\begin{equation}
O_l= V_l U_l^T,
\end{equation}
where 
\begin{equation}
B_l^* F_l = U_l \Sigma_l V_l^T
\end{equation}
is the singular value decomposition (SVD) of $B_l^* F_l$. Thus, we estimate $A_l$ by 
\begin{equation}
A_l = F_l V_l U_l^T.
\end{equation} 
In analogy with crystallography, the phase
information ($V_l U_l^T$) from the resolved homologous
structure appends the experimentally measured intensity information ($F_l$). We note that other magnitude correction schemes have been used in crystallography. For example, setting the magnitude to be twice the magnitude from the desired structure minus the magnitude from the known structure, has the desired effect of properly weighting the difference between the two structures, but also the undesired effect of doubling the noise level. The cryo-EM analog in this case would be estimating $A_l$ by 
\begin{equation}
A_l = 2F_l V_l U_l^T - B_l.
\end{equation}

\section{Orthogonal Replacement (OR)}
We move on to describe {\em Orthogonal Replacement}, our approach for resolving structures for which there does not exist a homologous structure.  Suppose $\phi^{(1)}$ and $\phi^{(2)}$
are two unknown structures
for which we have cryo-EM images.
We assume that their difference $\Delta \phi =\phi^{(2)} -\phi^{(1)}$ is known.
This can happen, for example, when
an antibody fragment of a known structure binds to a protein. We have two sets
of cryo-EM images, one from the protein alone, $\phi^{(1)}$
and another from the protein plus the antibody, $\phi^{(2)}$. Let $C_l^{(i)}$ be the matrices computed from the sample covariance
matrices of the 2D projection images of $\phi^{(i)}, (i=1,2)$.
Let $F_l^{(i)}$ be any matrix satisfying $C_l^{(i)} = F_l^{(i)}{F_l^{(i)}}^{*}$. We have
$A_l^{(i)} = F_l^{(i)} O_l^{(i)}$, where $O_l^{(i)} \in \text{O}(2l+1)$.
The matrices $O_l^{(i)}$ need to be determined for $i=1,2$ and $l=0,1,2, \cdots$. The
difference $A_l^{(2)} -A_l^{(1)}$ is known from
the 3D Fourier transform of the binding structure $\Delta \phi$. We have

\begin{equation}\label{lin}
A_l^{(2)}-A_l^{(1)}=F_l^{(2)}O_l^{(2)}-F_l^{(1)}O_l^{(1)}
\end{equation}

\subsection{Relaxation to a Semidefinite Program}
The least squares problem
\begin{equation}\label{noncon}
\min_{O_l^{(1)},O_l^{(2)} \in \text{O}(2l+1)}
\left\|A_l^{(2)}-A_l^{(1)}-F_l^{(2)}O_l^{(2)}+F_l^{(1)}O_l^{(1)} \right\|_F^2
\end{equation}
is a non-convex optimization problem with no closed form solution. We find
$O_l^{(1)}$ and $O_l^{(2)}$ using convex relaxation in the form of
semidefinite programming (SDP). We first homogenize (\ref{lin}) by introducing a
slack unitary variable $O_l^{(3)}$ and consider
the augmented linear system

\begin{equation}\label{aug}
(A_l^{(2)}-A_l^{(1)})O_l^{(3)}=F_l^{(2)}O_l^{(2)}-F_l^{(1)}O_l^{(1)}
\end{equation}
If the triplet $\{O_l^{(1)},O_l^{(2)},O_l^{(3)}\}$ is a solution to (\ref{aug}),
then the pair $\{O_l^{(1)}{O_l^{(3)}}^T, O_l^{(2)}{O_l^{(3)}}^T\}$
is a solution to the original linear system ($\ref{lin})$. The corresponding
least squares problem
\begin{equation}\label{sdp}
\min_{%
      \substack{%
        O_l^{(i)}\in \text{O}(2l+1) \\ i=1,2,3
}
      }
\left\|(A_l^{(2)}-A_l^{(1)})O_l^{(3)}-F_l^{(2)}O_l^{(2)}+F_l^{(1)}O_l^{(1)} \right\|_F^2
\end{equation}
is still non-convex. But it can be relaxed to an SDP. Let $Q \in
\mathbb{R}^{3(2l+1) \times 3(2l+1)}$ be a symmetric matrix, which can be
expressed as a $3 \times 3$ block matrix with block size $2l+1$, and the $ij$'th block is given by
\begin{equation}
\label{Q}
Q_{ij}=O_l^{(i)}{O_l^{(j)}}^T, \quad i,j=1,2,3
\end{equation}
It follows that $Q$ is positive semidefinite (denoted $Q \succeq 0$). Moreover, the three diagonal blocks of $Q$ are $Q_{ii}=I$ ($i=1,2,3$) and
$\text{rank}(Q)=2l+1$.
The cost
function in (\ref{sdp}) is quadratic in $O_l^{(i)} (i=1,2,3)$, so it is
linear in $Q$. The problem can be equivalently rewritten as
\begin{equation}
\min_Q \text{Tr}(WQ)
\end{equation}
over $Q \in \mathbb{R}^{3(2l+1) \times 3(2l+1)}$,
subject to $Q_{ii} = I$,  $\text{rank}(Q) = 2l+1$ and $Q \succeq 0$,
where the matrix $W$ can be written in terms of $A_l^{(2)}-A_l^{(1)}$,
$F_l^{(1)}$ and $F_l^{(2)}$. Here, we have only one non-convex constraint -- the
rank constraint. Upon dropping the rank constraint we arrive at an SDP that can be solved efficiently in
polynomial time in $l$. We extract
the orthogonal matrices
$O_l^{(i)}$ from the decomposition (\ref{Q}) of $Q$. If the
solution matrix $Q$ has rank greater than $2l+1$ (which is possible since we
dropped the rank constraint), then
we employ the rounding procedure of \cite{Cheeger}.  

\subsection{Exact Recovery and Resolution Limit}
We have the following theoretical guarantee on recovery of $O_l^{(1)}$ and $O_l^{(2)}$ using the SDP relaxation in the noiseless case:  
\begin{thm}\label{thm:recovery}
Assume that $A_l^{(1)}$ and $A_l^{(2)}\in\mathbb{R}^{K\times (2l+1)}$ are elementwise sampled from i.i.d. Gaussian $N(0,1)$, and $K>2l+1$, then the SDP method recovers $O_l^{(1)}$ and $O_l^{(2)}$ almost surely.
\end{thm}

The proof of Theorem~\ref{thm:recovery} is beyond the scope of this paper and is deferred to a separate publication. Theorem~\ref{thm:recovery} shows that the SDP method almost achieves the theoretical information limit, since by counting the degrees of freedom in (\ref{lin}) it is impossible to recover $O_l^{(1)}$ and $O_l^{(2)}$ if $K< 2l$. Indeed, the number of free parameters associated with an orthogonal matrix in
$\text{O}(2l+1)$ is $l(2l+1)$, while the number of equations in (\ref{lin}) is $K(2l+1)$. This introduces a natural resolution limit on structures that can be resolved. Only angular frequencies for which $l \leq \frac{K}{2}$ can be determined using OR.

\section{Numerical experiments}
\begin{figure}[t]
\begin{center}
\includegraphics[width=.99\columnwidth]{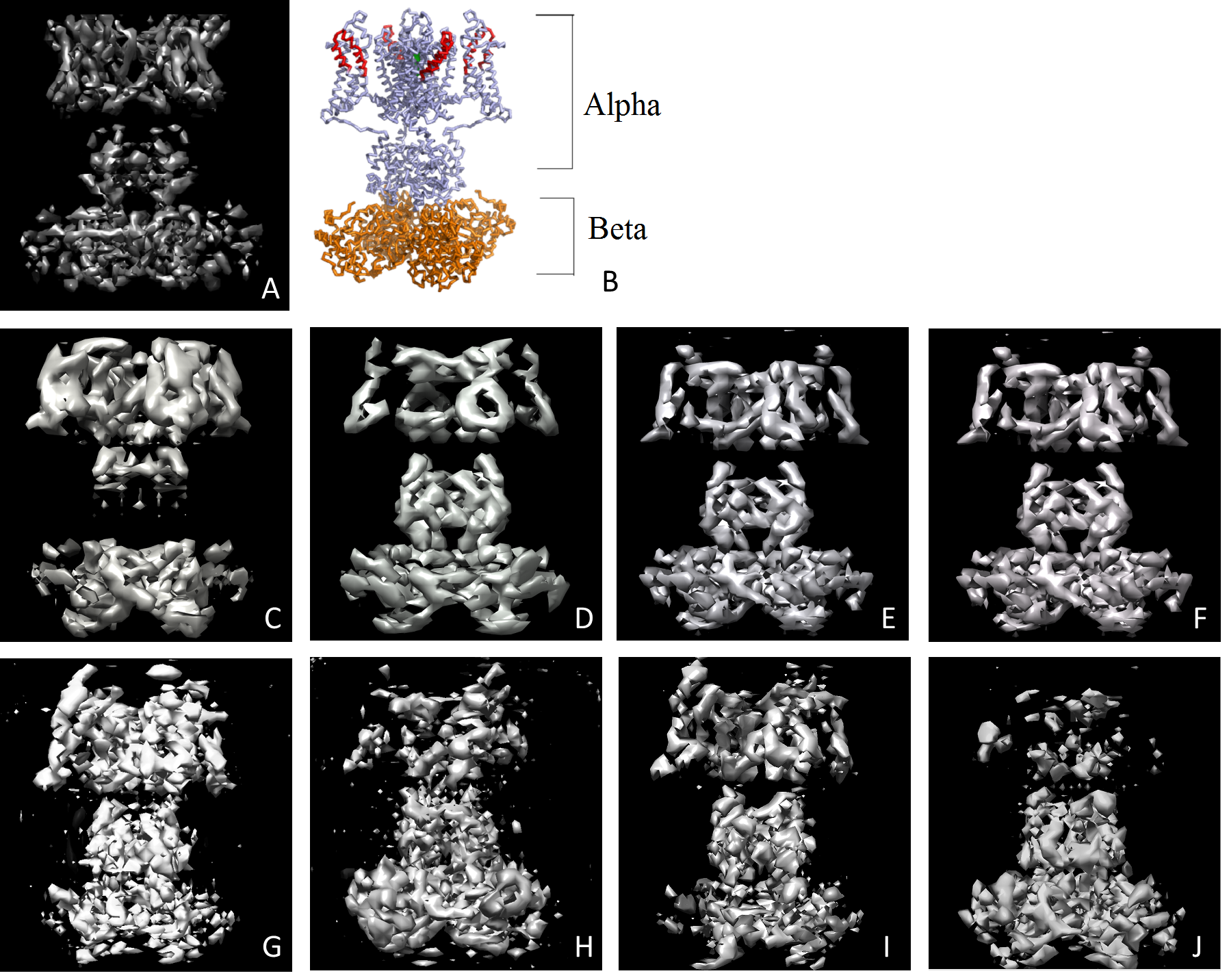}
\end{center}
\vspace{-.1in}
\caption{Kv1.2 potassium channel: A) Volume visualization in UCSF Chimera \cite{chimera}. B) Image from Protein Data Bank Japan (PDBj). C through F show reconstructions from clean images - C) OE with $\alpha_4$ known, D) OE with $\beta_4$ known, E) OR with $\alpha_4$ known, and F) OE with $\beta_4$ known. G through J show reconstructions from noisy images using OR - G) SNR=0.7 with $\alpha_4$ known, H) SNR=0.7 with $\beta_4$ known, I) SNR=0.35 with $\alpha_4$ known, and J) SNR=0.35 with $\beta_4$ known.}
\label{fig:truth}
\end{figure}

We present the results of numerical experiments on simulated images ($109\times 109$ pixels) of the Kv1.2 potassium
channel complex (Fig.~\ref{fig:truth} A and B) with clean and noisy
projection images. The experiments were performed in MATLAB in UNIX environment on an
Intel (R) Xeon(R) X7542 with 2 CPUs, having 6 cores each, running at 2.67 GHz, and with
256 GB RAM in total. To solve the SDP we used the MATLAB package CVX \cite{cvx}, and to compute the covariance matrix of the 2D images we used the steerable PCA procedure \cite{Zhao1}.

Kv1.2 is a dumbbell-shaped particle consisting of two subunits - a small $\beta_4$ subunit
and a larger $\alpha_4$ subunit, connected by a central connector. We performed
experiments using OE and OR, assuming one of the subunits (e.g., $\alpha_4$) is known,
while the other is unknown. In the case of OR, we additionally used projection
images of the unknown subunit. 

\begin{figure}[]
\centering
  \includegraphics[width=.9\columnwidth]{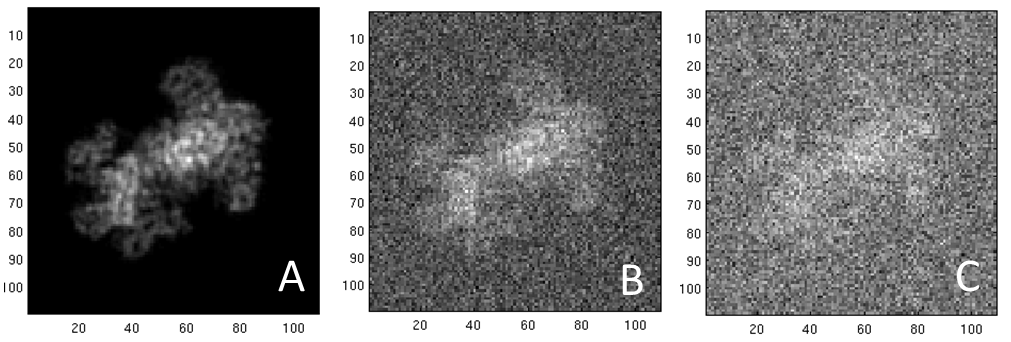}
  \caption{Projection images at different values of SNR: A) Clean image, B) SNR=0.7, and
  C) SNR=0.35.}
\label{fig:proj}
\end{figure}

\subsection{Clean and Noisy Projections}
We reconstruct the structure from both clean and noisy projection images. The
reconstruction of Kv1.2 obtained from clean images using OE and OR is shown in
Fig.~\ref{fig:truth} C through F. We used the true $C_l$ matrices for the known subunit, and a maximum
$l$ of 30. We tested OR to reconstruct Kv1.2 from noisy projections at various values of
SNR. A sample projection image at different values of SNR is shown in Fig.~\ref{fig:proj}.
The $C_l$ matrices were estimated from the noisy projection images. In
Fig.~\ref{fig:truth} G through J we show the reconstructions obtained from 10000 projections using OR
at SNR=0.7, and from 40000 projections using OR at SNR=0.35. In our simulations with 10000 images, it takes $416$ seconds to perform steerable PCA, $194$ seconds to calculate the $C_l$ matrices using the maximum $l$ as 30, and the time to solve the SDP as a function of $l$ ranges from 5 seconds for $l=5$ to $194$ seconds for $l=30$.

\subsection{Comparison between OE and OR}
\begin{figure}[t]
  \centering
  \includegraphics[scale=0.25]{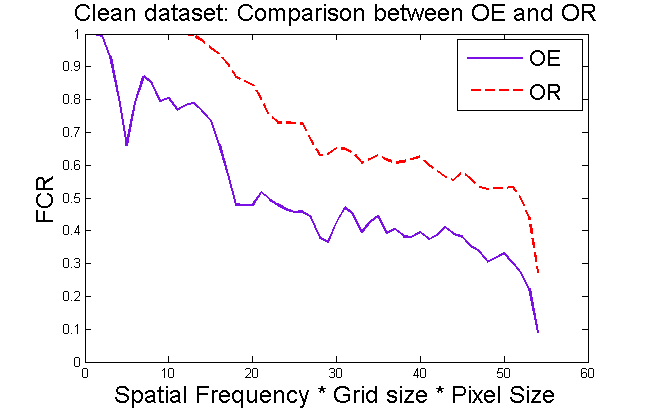}
  \caption{FCR curve for reconstruction from $\beta_4$ (clean images)}\label{fig:fsc}
\end{figure}

We quantify the `goodness' of the reconstruction using the Fourier Cross Resolution (FCR)
\cite{fcr}. In Fig. \ref{fig:fsc} we show the FCR curves for the
reconstruction from the $\beta_4$ complex using OE and OR. The additional information in
OR, from the projection images of $\alpha_4$, results in a better reconstruction, as seen
from the FCR curve. The Kv1.2 complex has C4 symmetry, which reduces the rank of the
$C_l$ matrices. Our experiment thus benefits from the reduced size of the orthogonal
matrices to be recovered. 

\section{Summary}
We presented two new approaches based on Kam's theory for {\em ab-initio} modelling of macromolecules for SPR from cryo-EM.  
Ab-initio modelling of small complexes is a challenging problem in cryo-EM because it is difficult to detect common lines between noisy projection images at low SNR. 
Our methods only require reliable estimation of the covariance matrix of the projection images which can be met even at low SNR if the number of images is sufficiently large.    
In future work we plan to estimate the covariance matrix when images are not centered, and to apply our methods to experimental datasets.

\bibliographystyle{IEEEbib}
\bibliography{ur_amit}

\begin{thebibliography}{10}

\bibitem{Frank1}
J.~Frank,
\newblock {\em Three-Dimensional Electron Microscopy of Macromolecular
  Assemblies : Visualization of Biological Molecules in Their Native State:
  Visualization of Biological Molecules in Their Native State},
\newblock Oxford University Press, USA, 2006.

\bibitem{resolution_revolution}
W.~K\"{u}hlbrandt,
\newblock ``The resolution revolution,''
\newblock {\em Science}, vol. 343, pp. 1443--1444, 2014.

\bibitem{Revol}
X.~Bai, G.~McMullan, and S.H.W Scheres,
\newblock ``How cryo-em is revolutionizing structural biology,''
\newblock {\em Trends in Biochemical Sciences}, in press.

\bibitem{wang06}
L.~Wang and F.~J. Sigworth,
\newblock ``Cryo-{EM} and single particles,''
\newblock {\em Physiology (Bethesda)}, vol. 21, pp. 13--18, 2006.

\bibitem{vanheel00}
M.~van Heel, B.~Gowen, R.~Matadeen, E.~V. Orlova, R.~Finn, T.~Pape, D.~Cohen,
  H.~Stark, R.~Schmidt, and A.~Patwardhan,
\newblock ``Single particle electron cryo-microscopy: {T}owards atomic
  resolution,''
\newblock {\em Q. Rev. Biophys.}, vol. 33, pp. 307--369, 2000.

\bibitem{gridding}
P.~Penczek, R.~Renka, and H.~Schomberg,
\newblock ``Gridding-based direct {F}ourier inversion of the three-dimensional
  ray transform,''
\newblock {\em J. Opt. Soc. Am. A}, vol. 21, pp. 499--509, 2004.

\bibitem{Radermacher}
A.~Verschoor M.~Radermacher, T.~Wagenknecht and J.~Frank,
\newblock ``Three-dimensional reconstruction from a single-exposure, random
  conical tilt series applied to the 50s ribosomal subunit of escherichia
  coli,''
\newblock {\em Journal of Microscopy}, vol. 146, no. 2, pp. 113--136, 1987.

\bibitem{Radermacher2}
A.~Verschoor M.~Radermacher, T.~Wagenknecht and J.~Frank,
\newblock ``{Three-dimensional structure of the large ribosomal subunit from
  Escherichia coli},''
\newblock {\em EMBO J}, vol. 6, no. 4, pp. 1107--14, 1987.

\bibitem{Goncharov}
A.B. Goncharov,
\newblock ``Integral geometry and three-dimensional reconstruction of randomly
  oriented identical particles from their electron microphotos,''
\newblock {\em Acta Applicandae Mathematicae}, vol. 11, pp. 199--211, 1988.

\bibitem{Salzman}
D.B. Salzman,
\newblock ``A method of general moments for orienting {2D} projections of
  unknown {3D} objects,''
\newblock {\em Comput. Vision Graph. Image Process.}, vol. 50, no. 2, pp.
  129--156, May 1990.

\bibitem{Goncharov1986}
B.K. Vainstein and A.B. Goncharov,
\newblock ``Determining the spatial orientation of arbitrarily arranged
  particles given their projections,''
\newblock {\em Dokl. Acad. Sci. USSR}, vol. 287, no. 5, pp. 1131--1134, 1986,
\newblock English translation: Soviet Physics Doklady, Vol. 31, p.278.

\bibitem{vanHeel1987}
M.~Van~Heel,
\newblock ``Angular reconstitution: a posteriori assignment of projection
  directions for 3d reconstruction,''
\newblock {\em Ultramicroscopy}, vol. 21, no. 2, pp. 111--123, 1987.

\bibitem{SS_11}
A.~Singer and Y.~Shkolnisky,
\newblock ``Three-dimensional structure determination from common lines in
  cryo-{EM} by eigenvectors and semidefinite programming,''
\newblock {\em SIAM Journal on Imaging Sciences}, vol. 4, pp. 543--572, 2011.

\bibitem{Zhao}
Z.~Zhao and A.~Singer,
\newblock ``Rotationally invariant image representation for viewing direction
  classification in cryo-{EM},''
\newblock {\em Journal of Structural Biology}, vol. 186, no. 1, pp. 153 -- 166,
  2014.

\bibitem{kam1980}
Z.~Kam,
\newblock ``The reconstruction of structure from electron micrographs of
  randomly oriented particles,''
\newblock {\em Journal of Theoretical Biology}, vol. 82, no. 1, pp. 15 -- 39,
  1980.

\bibitem{Natr2001a}
F.~Natterer,
\newblock {\em The Mathematics of Computerized Tomography},
\newblock Classics in Applied Mathematics. SIAM: Society for Industrial and
  Applied Mathematics, 2001.

\bibitem{keller}
J.~B. Keller,
\newblock ``Closest unitary, orthogonal and {Hermitian} operators to a given
  operator,''
\newblock {\em Mathematics Magazine}, vol. 48, no. 4, pp. 192--197, 1975.

\bibitem{Cheeger}
A.~S. Bandeira, A.~Singer, and D.~A. Spielman,
\newblock ``A {Cheeger} inequality for the graph connection {Laplacian},''
\newblock {\em SIAM Journal on Matrix Analysis and Applications}, vol. 34, no.
  4, pp. 1611--1630, 2013.

\bibitem{cvx}
M.~Grant and S.~Boyd,
\newblock ``{CVX}: Matlab software for disciplined convex programming, version
  2.1,'' \url{http://cvxr.com/cvx}, Mar. 2014.

\bibitem{Zhao1}
Z.~Zhao and A.~Singer,
\newblock ``Fourier {Bessel} rotational invariant eigenimages,''
\newblock {\em J. Opt. Soc. Am. A}, vol. 30, no. 5, pp. 871--877, May 2013.

\bibitem{chimera}
E.~F. Pettersen, T.~D. Goddard, C.~C. Huang, G.~S. Couch, D.~M. Greenblatt,
  E.~C. Meng, and T.~E. Ferrin,
\newblock ``{UCSF Chimera--}a visualization system for exploratory research and
  analysis,''
\newblock {\em Journal of Computational Chemistry}, vol. 25, no. 13, pp.
  1605--1612, 2004.

\bibitem{fcr}
P.~A. Penczek,
\newblock ``Chapter three - resolution measures in molecular electron
  microscopy,''
\newblock in {\em Cryo-EM, Part B: 3-D Reconstruction}, Grant~J. Jensen, Ed.,
  vol. 482 of {\em Methods in Enzymology}, pp. 73 -- 100. Academic Press, 2010.

\end{thebibliography}

\end{document}